\RequirePackage{fix-cm}
\documentclass[smallextended]{svjour3}       
\smartqed  
\usepackage{graphicx,amsmath,geometry,subcaption,enumitem,epstopdf,hyperref,placeins}
\usepackage{eso-pic}
\newcommand\AtPageUpperMyright[1]{\AtPageUpperLeft{%
 \put(\LenToUnit{0.5\paperwidth},\LenToUnit{-1cm}){%
     \parbox{0.5\textwidth}{\raggedleft\fontsize{9}{11}\selectfont #1}}%
 }}%
\newcommand{\conf}[1]{%
\AddToShipoutPictureBG*{%
\AtPageUpperMyright{#1}
}
}

\begin{document}

\title{Multi Agent Collaboration for Building Construction}
\subtitle{}
\author{ Kumar Ankit$^{*}$ \and Lima Agnel Tony$^{*}$ \and Shuvrangshu Jana$^{\#}$  \and Debasish Ghose$^{\dagger}$}


\institute{ $^{*}$ Doctoral Research Scholar,$^{\#}$ Post Doctoral Fellow, $^{\dagger}$Professor,\\ 
            Guidance Control and Decision Systems Laboratory \\
              Department of Aerospace Engineering, Indian institute of Science,  Bangalore-12, India\\
              \email{kumarankit@iisc.ac.in}  of F. Author  %
              \email{limatony@iisc.ac.in} of S. Author
              \email{shuvrangshuj@iisc.ac.in} of T. Author
              \email{dghose@iisc.ac.in} of F. Author
}

\date{}

\maketitle
\conf{MBZIRC Symposium\\ADNEC, Abu Dhabi, UAE\\ 26-27 February 2020}
\begin{abstract}
This paper details the algorithms involved and task planner for vehicle collaboration in building a structure. This is the problem defined in the challenge 2 of Mohammed Bin Zayed International Robotic Challenge 2020 (MBZIRC). The work addresses various aspects of the challenge for Unmanned Aerial Vehicles (UAVs) and Unmanned Ground Vehicle (UGV). Challenge involves repeated pick and place operations using UAVs and UGV to build two structures of different shape and sizes. The algorithms are implemented using Robot Operating System (ROS) frame work and visualised in Gazebo. The whole developed architecture could readily be implemented in suitable hardware.
\end{abstract}
\section{Introduction}
\label{intro}
Autonomy has undergone major changes with the advent of Unmanned Aerial Vehicles (UAVs), especially in the field of autonomy.  UAVs are being employed for diverse applications thus reducing human efforts and increasing efficiency. Mohammed Bin Zayed International Robotic Challenge 2020 has defined challenging problems to solve which requires the right proportion of hardware design and software edge to achieve good results. 

Aerial pick and place has been addressed in literature using different methodologies. A model predictive control based approach is used in \cite{rf1} to achieve grabbing task for a quadrotor with a two link manipulator. A multi-manipulator coordination for pick and place operation for product assembly is discussed in \cite{rf2} which is robust against pattern variations at the pick up end. A stereo vision based approach is followed in \cite{rf3} where a remotely located industrial robot is manipulated via LAN. A vision based manipulation is described in \cite{rf4} where a humanoid identifies, picks and places surgical tools in their respective trays. A fast pick and place operation of spherical objects is discussed in \cite{rf5} while detection, localisation and motion planning for pick and place of objects on a conveyor belt is developed and implemented in \cite{rf6}. In \cite{rf7}, a multi-degree of freedom manipulator is developed which is controlled using classical controller and is employed for diverse tasks like pick and place, insertion and valve turning. Pick and place operation of deformable objects is carried out in \cite{rf8} where point cloud is used to determine the feasibility of operation. The controller design of a manipulator for indoor application is developed in \cite{rf9} where, the applicability of the manipulator is pick and place operation of unknown mass. Load transportation using aerial robots and cooperative operations between aerial vehicles and aerial and ground vehicles is surveyed in \cite{rf10} which analyzes various kind of loads that could be transported and the stability of the resultant systems.

Challenge 2 is one such scenario which requires UAVs and Unmanned Ground Vehicle (UGV) to build a wall of given sequence. This involves the vehicles to autonomously search, detect, pick up and place according to a given sequence. Given the specifications of the bricks to be used to achieve the task, specially designed robotic manipulators and algorithms are needed to successfully execute the challenge. In this paper, the software aspects of the challenge is taken care of, where a single degree of freedom manipulator is employed to complete the challenge. Multiple aspects including the vision, manipulation, path planning, multi-vehicle coordination and task assignment are also addressed in this work.

The paper is organised as: Section \label{S2} details the problem of building construction using UAVs and UGV, as per the challenge specifications of MBZIRC 20. Section \label{S3} details the set up and architecture developed in Robot Operating System (ROS) to address the problem. Section \label{S4} lays out the process from a UAV perspective while Section \ref{S5} gives the same in UGV perspective. Section \ref{S6} details the blanket layer of the entire system which takes care of multiple vehicles' task allocation and planning and other relevant aspects of the whole. Section \ref{S7} gives out simulation results and Section \ref{S8} concludes the paper.
\section{Problem statement}\label{S2}
\begin{figure}
    \centering
\begin{subfigure}{0.32\textwidth}
\includegraphics[width=0.9\linewidth, height=3.5cm]{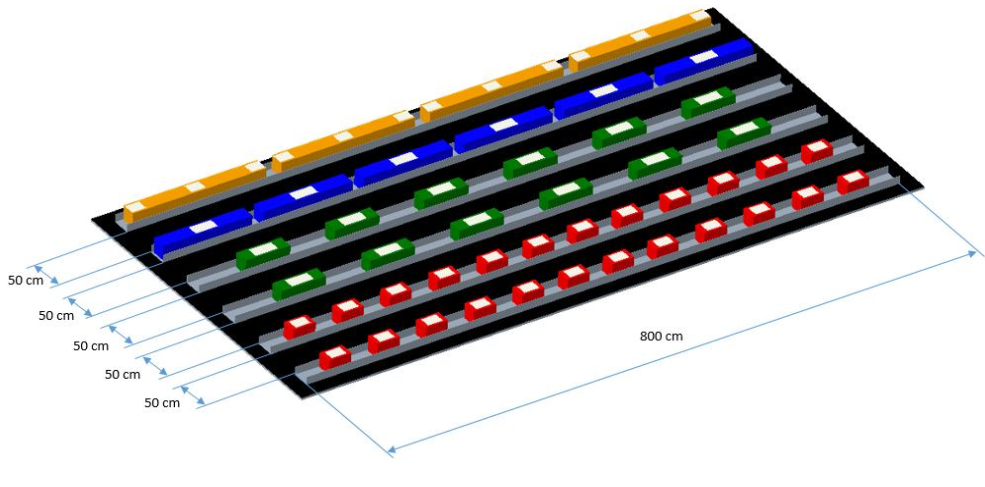} 
\caption{UAV brick stack}
\end{subfigure}
\begin{subfigure}{0.32\textwidth}
\includegraphics[width=0.9\linewidth, height=3.5cm]{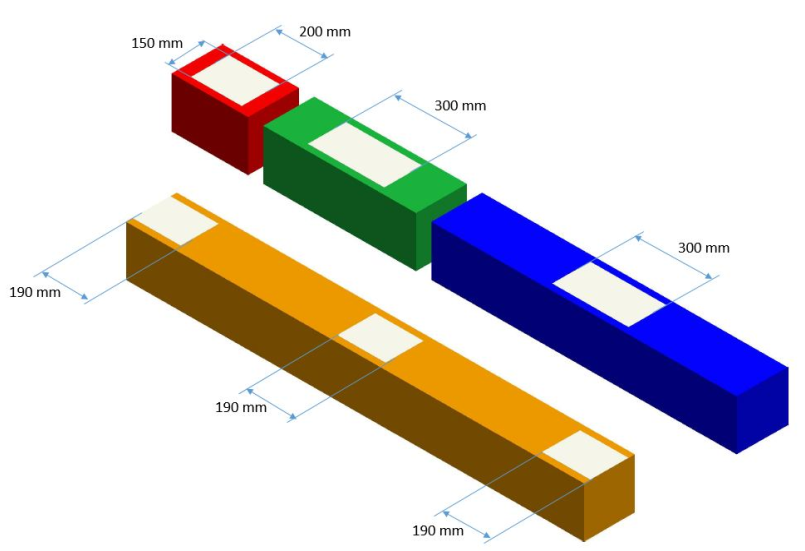} 
\caption{Bricks with white labels marking the magnetic portion}
\end{subfigure}
\begin{subfigure}{0.32\textwidth}
\includegraphics[width=0.9\linewidth, height=3.5cm]{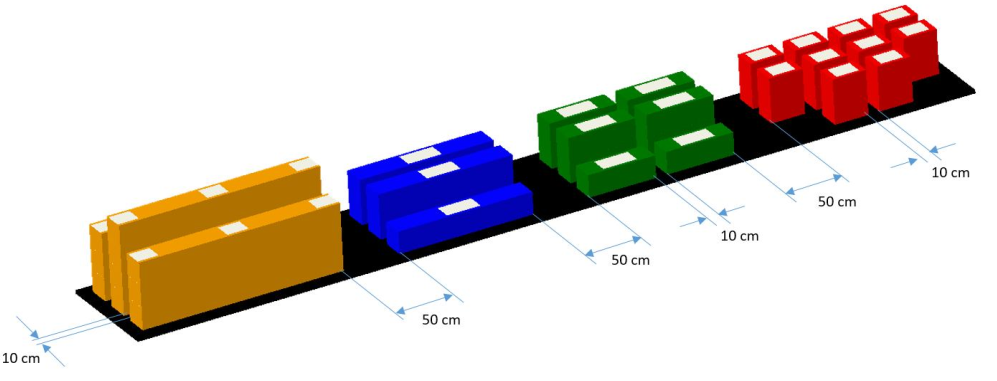} 
\caption{UGV brick stack}
\end{subfigure}
\caption{Brick piles}
\label{f1}
\end{figure}

\begin{figure}
    \centering
\begin{subfigure}{0.32\textwidth}
\includegraphics[width=0.9\linewidth, height=3.5cm]{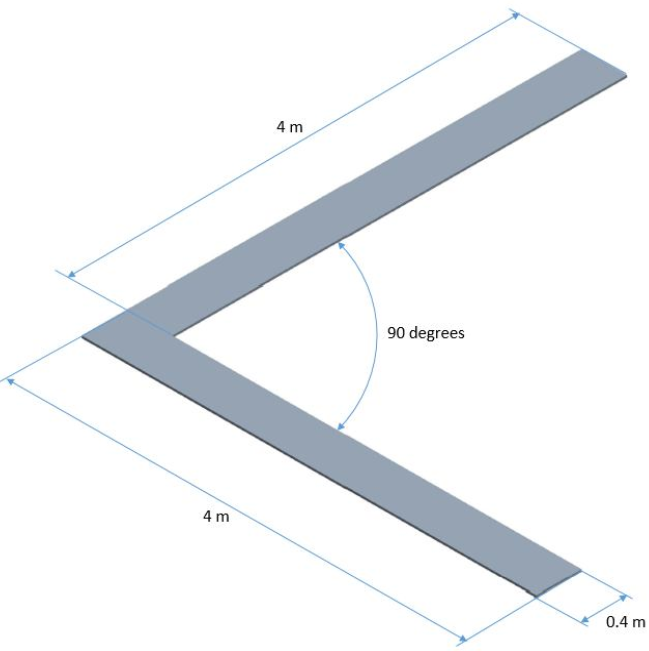} 
\caption{UGV construction site}
\end{subfigure}
\begin{subfigure}{0.32\textwidth}
\includegraphics[width=0.9\linewidth, height=3.5cm]{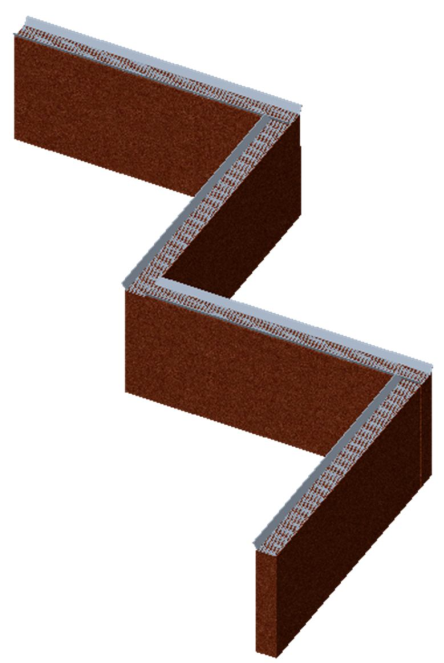} 
\caption{UAV construction site}
\end{subfigure}
\begin{subfigure}{0.32\textwidth}
\includegraphics[width=1\linewidth, height=3.5cm]{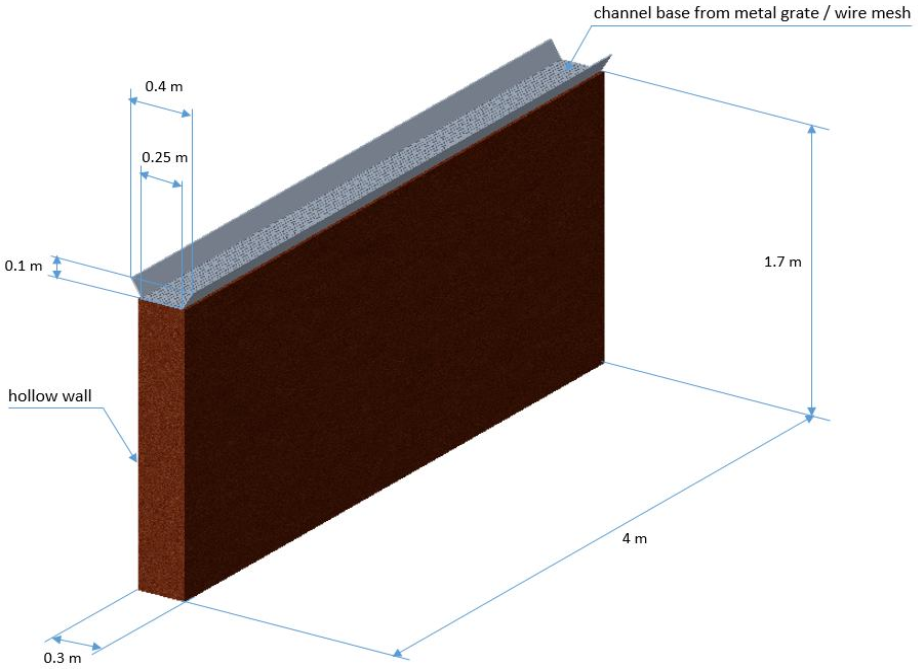} 
\caption{UAV channel unit specification}
\end{subfigure}
\caption{Construction site for the vehicles}
\label{f2}
\end{figure}

Three UAVs and UGV collaborate to build separate walls in an arena is of size 50 m x 40 m x 20 m. There are four different type of bricks, varying in dimension and color, as given in Table \ref{tab1}.
\begin{table}[t]
\centering
    \begin{tabular}{l|c|c}
    \hline
        Brick &Dimension (cm)& Weight (kg) \\
        \hline
        Red & 30 x 20 x 20 & 1\\
        Green & 60 x 20 x 20 & 1\\
        Blue & 120 x 20 x 20 & 1.5\\
        Orange & 180 x 20 x 20& 2\\
        \hline
    \end{tabular}
    \caption{Brick specifications}
    \label{tab1}
\end{table}
The bricks are separately stacked for UAVs and UGV. The UAV stacks are distributed to aid easy picking while the UGV stacks are uniform, as shown in Fig. \ref{f1}.
There is dedicated construction site for the UAVs and the UGV. The UGV has to build the wall in the shape of `L' which has an arm length of 4 m and a height of 1 m. UAVs have a zig zag construction site as shown in Fig. \ref{f2}. The construction platform has a height of 1.7 m and has channel on top of it. The channel base is made of wire mesh. This reduces the destabilizing effect of the UAV downwash while placing over the channel.
UAVs should build two layers of bricks over this channel. For UGV and UAVs, one 4 m channel is reserved for the orange bricks. This paper details the algorithms and packages developed for executing the entire challenge 2 in Robot Operating System (ROS) framework. The maximum speed of UAVs are restricted to 15 kmph and that of UGV is restricted to 30 kmph due to safety considerations.
\section{ROS architecture}\label{S3}
The simulations are visualised using Gazebo 9.11. Following environment is set up to meet the competition specifications.
\subsection{Arena}
\begin{figure}
    \centering
    \includegraphics[scale=0.25]{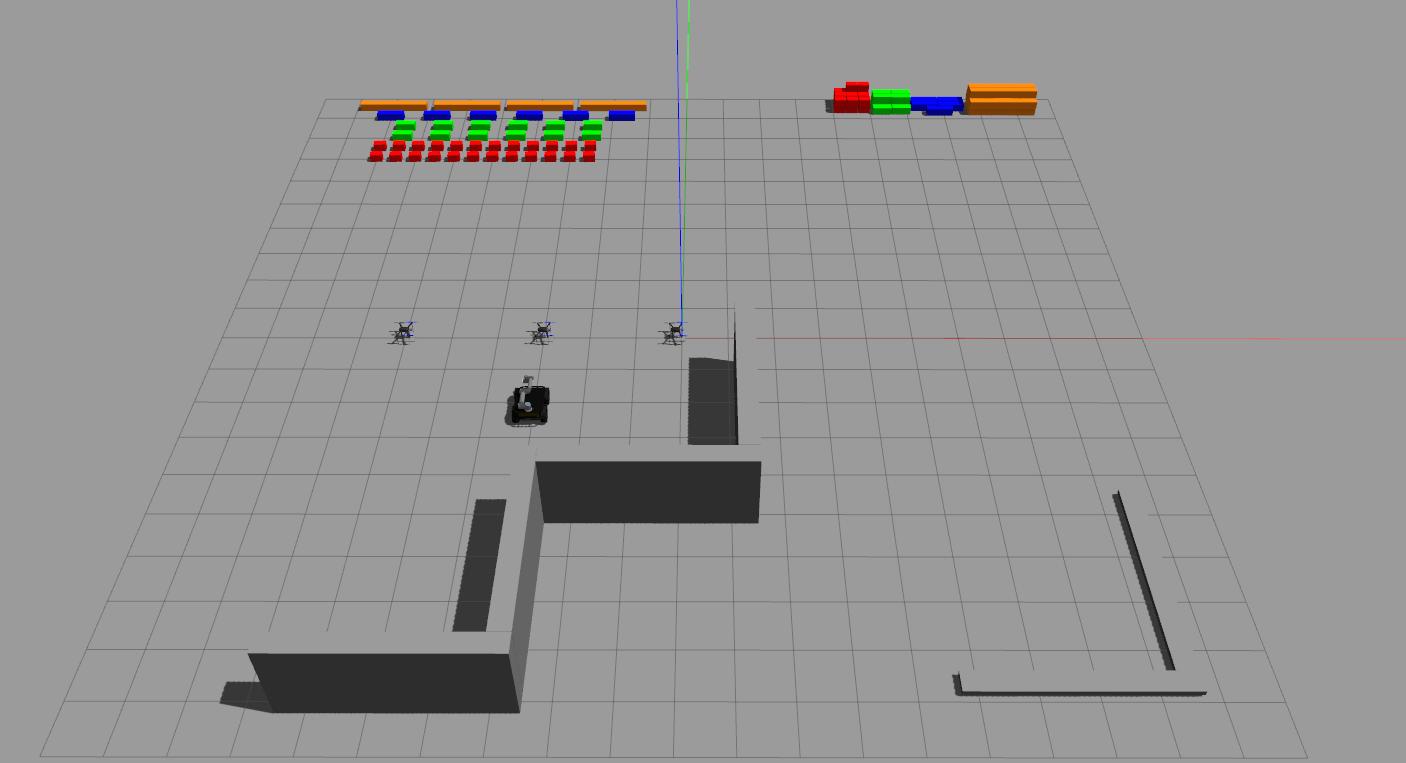}
    \caption{Challenge 2 arena in Gazebo}
    \label{f3}
\end{figure}

An arena is created in Gazebo with similar construction sites as specified by the competition organisers. Three UAVs and one UGV is deployed in this environment. The arena is as shown in Fig. \ref{f3}.
\subsection{UGV and UAVs}
For simulating the UGV, model of Husky by Clearpathrobotics is ported into the Gazebo environment. This is integrated with a UR5 manipulator. The UR5 has a reach radius of 850 mm and payload of up to 5 kg. Two monocular cameras are employed in the UGV, one forward facing and another downward facing. The forward facing one aids in search and exploration while the downward facing camera on the manipulator is employed for pick and place operations.

\begin{figure}
    \centering
\begin{subfigure}{0.32\textwidth}
\includegraphics[width=0.9\linewidth, height=3.5cm]{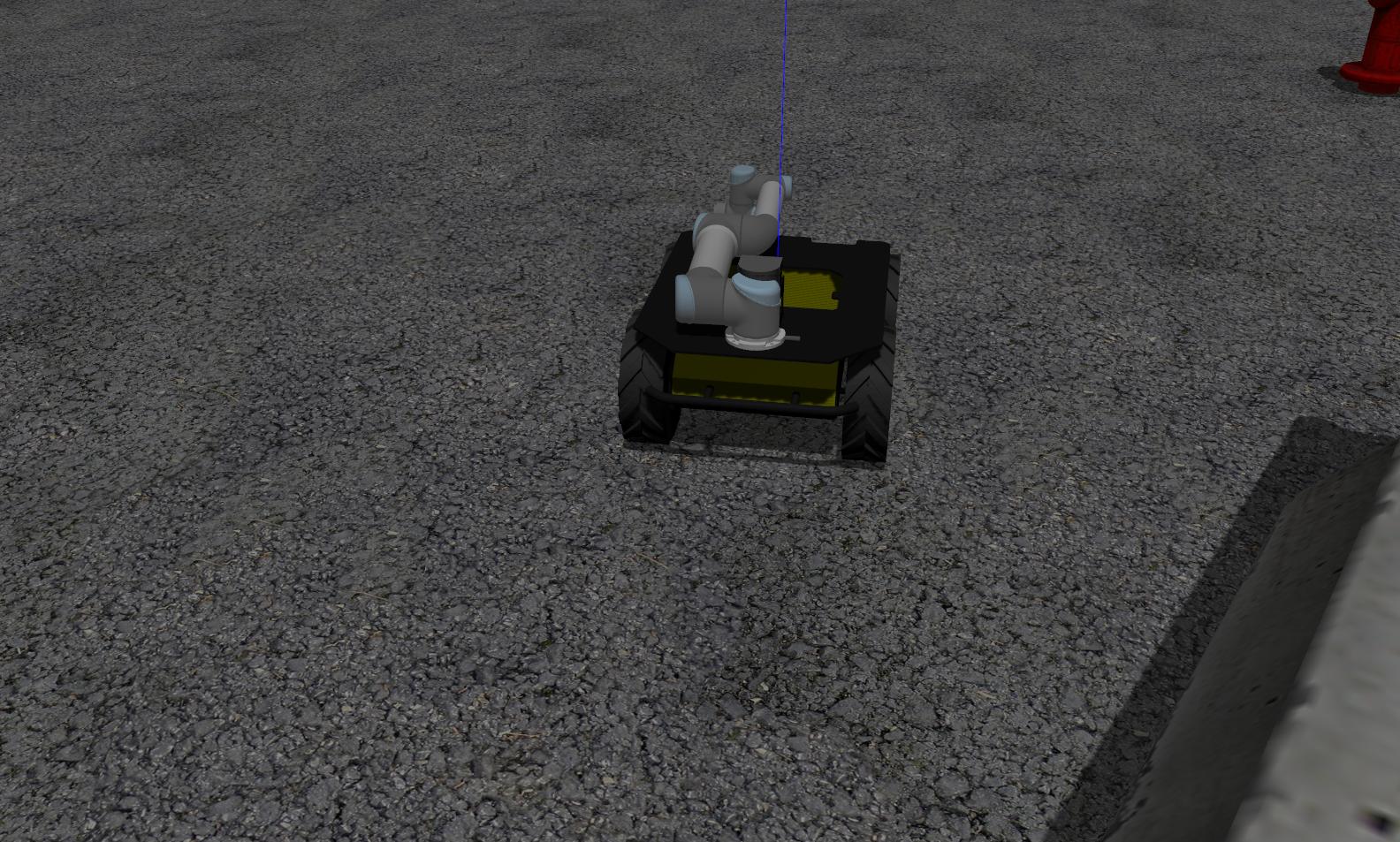} 
\caption{Husky UGV model with UR5 manipulator}
\end{subfigure}
\begin{subfigure}{0.32\textwidth}
\includegraphics[width=0.9\linewidth, height=3.5cm]{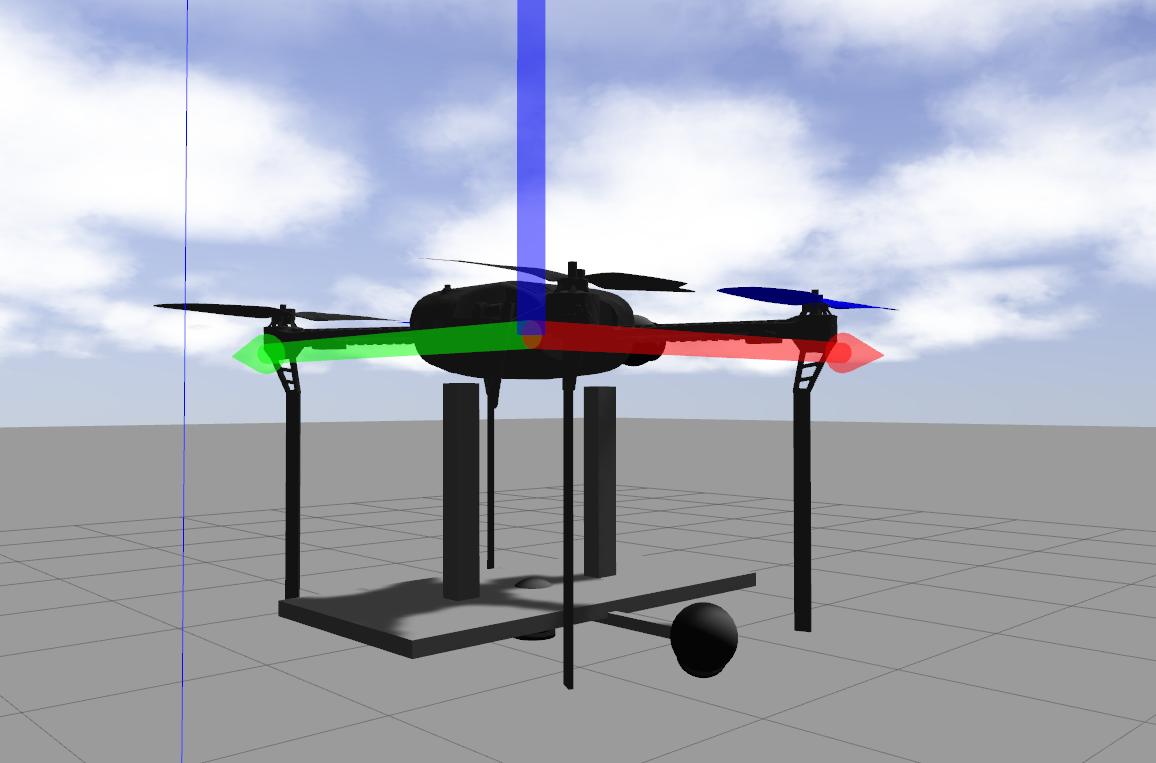} 
\caption{Iris drone with downward facing camera}
\end{subfigure}
\begin{subfigure}{0.32\textwidth}
\includegraphics[width=1\linewidth, height=3.5cm]{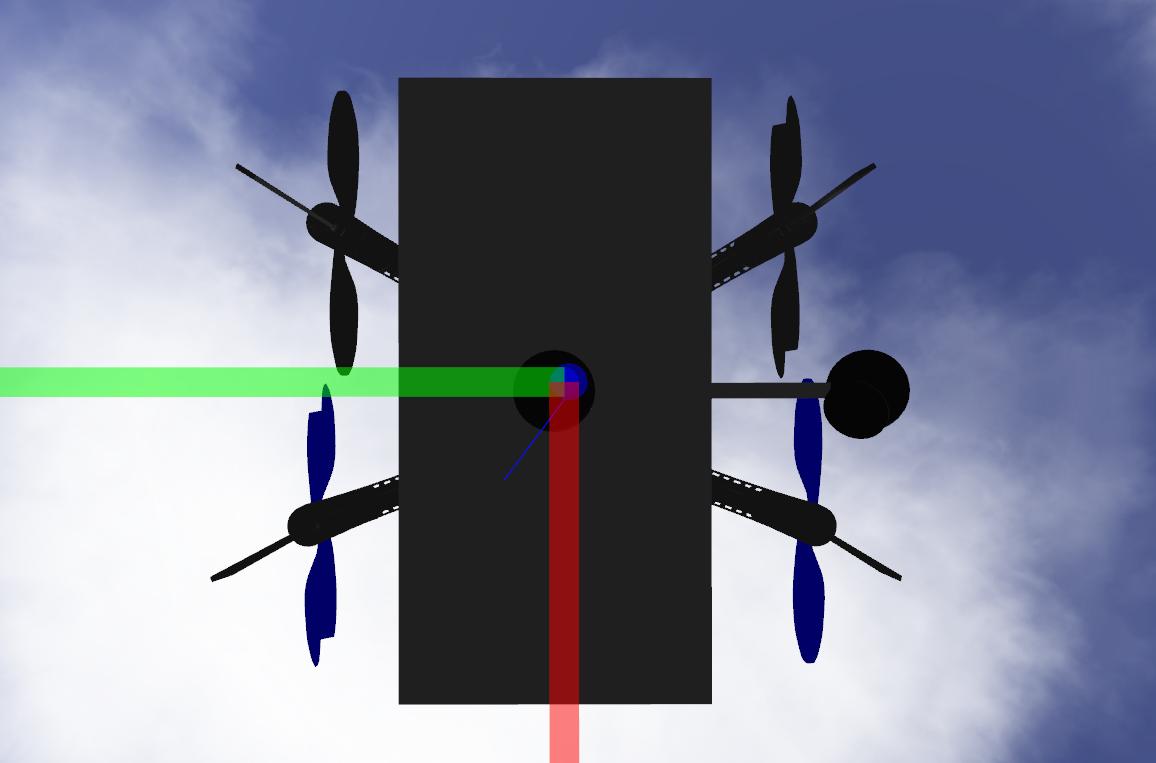} 
\caption{Iris drone with manipulator }
\end{subfigure}
    \caption{Vehicle models used for simulation}
    \label{f4}
\end{figure}

The UAVs use IRIS drone model. It's a quad rotor and is equipped with downward facing manipulator with electro-permanent gripping mechanism. Two monocular cameras are employed in the UAV, one on the manipulator and another to the side. The camera on manipulator aids pick operation while the sideways one helps in place operations. The UGV and UAV models are as in Fig \ref{f4}. 
\section{Wall building using UAV}\label{S4}
The following sub tasks are involved in autonomous wall construction by a UAV.
\subsection{Arena exploration}
Search is carried out to detect the locations of brick piles as well as the construction site for the vehicles. For this, a downward facing camera is employed. The drone does exploration using a lawn-mover pattern to identify the pile locations and construction site locations for both UAV and UGV. This information is then shared to the master through ROS interface.
\subsection{Pick and place operation}
\begin{figure}
    \centering
\begin{subfigure}{0.45\textwidth}
\includegraphics[scale=0.775]{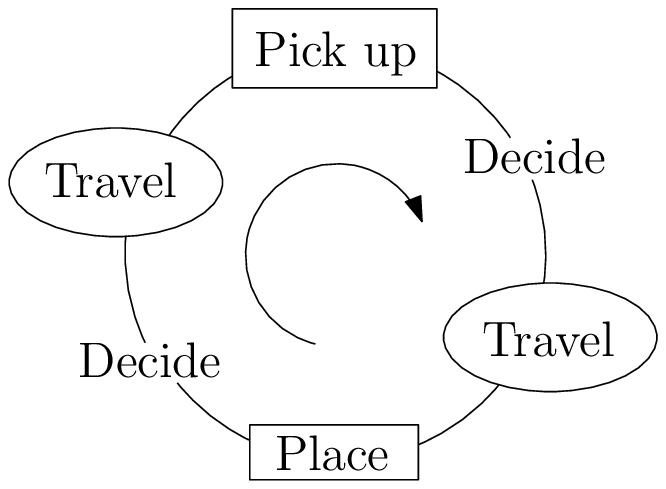} 
\caption{}
\end{subfigure}
\begin{subfigure}{0.45\textwidth}
\includegraphics[scale=0.155]{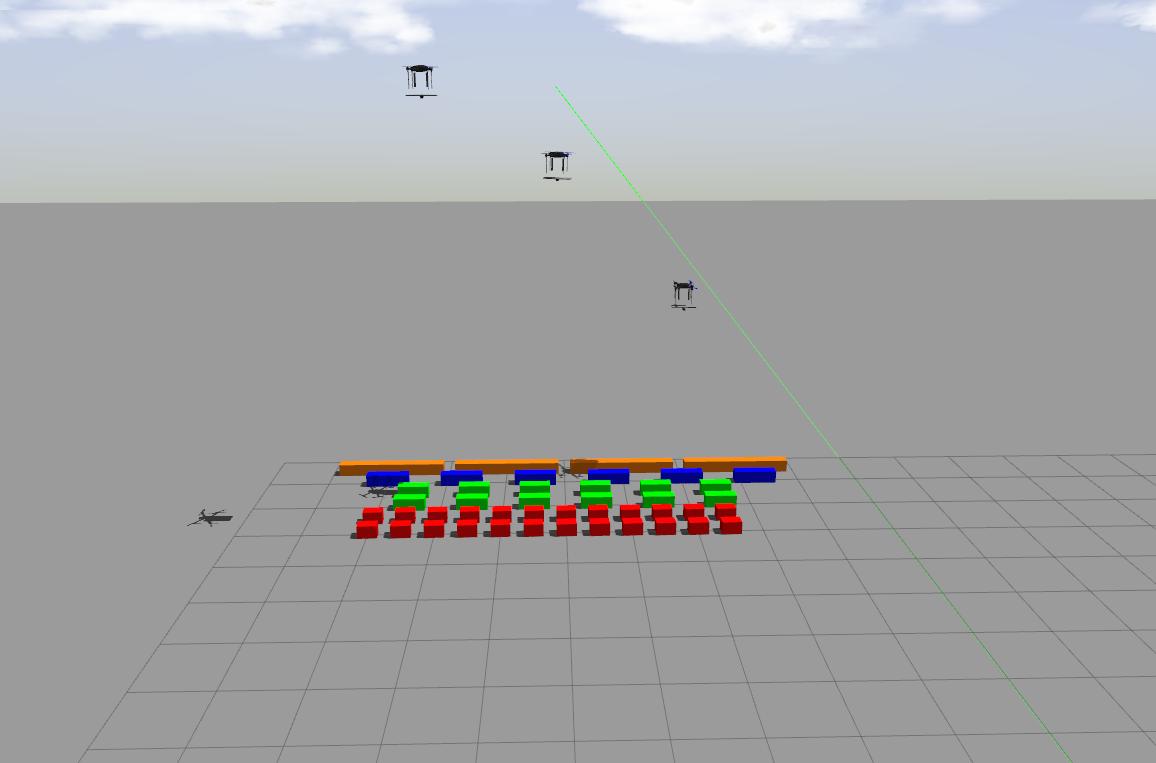} 
\caption{}
\end{subfigure}
\caption{(a) The sequence of operation in pick and place (b) UAVs with at different assigned corridors }
\label{f5}
\end{figure}

Challenge two requires repetitive motion of the vehicles between the brick stack and the construction site with minimal interference with the paths of each other, as represented as in Fig. \ref{f5}(a).
The two `decisions` blocks in Fig. \ref{f5}(a) represent the process of choosing the placing location and pick up channel for pick and place tasks, respectively. This involves dedicated path assignment for the vehicles to optimise performance. The operation of pick and place involves the following sub operations.
\subsubsection{Path planning}
Using the information from exploration, the UAV should traverse from the pile location to the construction site. For this, a path is planned joining the locations. UAVs are assigned corridors at different altitudes such that the vehicles won't collide with each other, as shown in Fig. \ref{f5}(b).
\subsubsection{Vision based brick detection and alignment and pick up}
Once the vehicles reach the location of pile, fine corrections are required in its position and orientation so as to pick up the brick. For this, vision is employed. The feed from the downward facing camera is processed to detect the brick. The detector uses segmentation to detect the white patch on the brick. With the contours obtained, following information is determined.
\begin{itemize}
\item Brick Center: This is computed based on the average value of the contour points.
\item Yaw: An ellipse is fitted to the polygon made from the contour points and the angle of the major axis in image plane is taken as desired yaw for the vehicle.
\item Area: The total area under the polygon is taken and used as a depth parameter while descending to pickup.
\end{itemize}
Alignment is based on the center obtained using image processing. The drone positions above the center of the brick by the following steps.
\begin{itemize}
\item Error is computed between camera center and brick center along both $x$ and $y$ axes as
    \[e_{Cx} = (b_{Cx} - w_{img}/2)\]
    \[e_{Cy} = (b_{Cy} - h_{img}/2)\]
where, $e_{Cx}$ and $e_{Cy}$ are computed error values, $b_{Cx}$ and $b_{Cy}$ correspond to the center of detected brick and $w_{img}$ and $h_{img}$ are the width and height of the image captured.
\item Desired velocities in X and Y axis are then obtained using a PD controller over the error in both the axes.
    \[V_x = kp_{Cx}*e_{Cx} + kd_{Cx}*(\Delta e_{Cx})\]
    \[V_y = kp_{Cy}*e_{Cy} + kd_{Cy}*(\Delta e_{Cy})\]
    where, $kp$ and $kd$ are the respective proportional and derivative gains.
\end{itemize}
When centered, the UAV yaws to align the brick along the manipulator using the yaw information obtained from image processing. 
    \[\omega_z = kp_{yaw}*b_{yaw} + kd_{yaw}*(\Delta b_{yaw})\]
where, $\omega_z$ is the angular velocity of the UAV in vertical axis and $b_{yaw}$ is the change in yaw from desired and current yaw values. After aligning, UAV descends to touch down on the brick. Descending velocity is calculated by deploying a PD controller over the error between desired and current area spanned by the brick in the image plane. 
    \[e_{area} = b_{area} - d_{area}\]
where, $d_{area}$ is desired area and $b_{area}$ is the instantaneous brick area visible to the camera. The descend velocity is computed as
    \[V_z = kp_{area}*e_{area} + kd_{area}*(\Delta e_{area})\]
While descending, the UAV may drift away from brick center or misalign. To tackle this the centering and aligning check is carried out at each iteration of the PD loop. After reaching the proper height, the Electro-Permanent Magnet is activated to attach the brick to the manipulator. After confirming the attachment, the drone lifts off to its selected corridor height.
\begin{figure}
\centering
\begin{subfigure}{0.3\textwidth}
\includegraphics[scale=0.1]{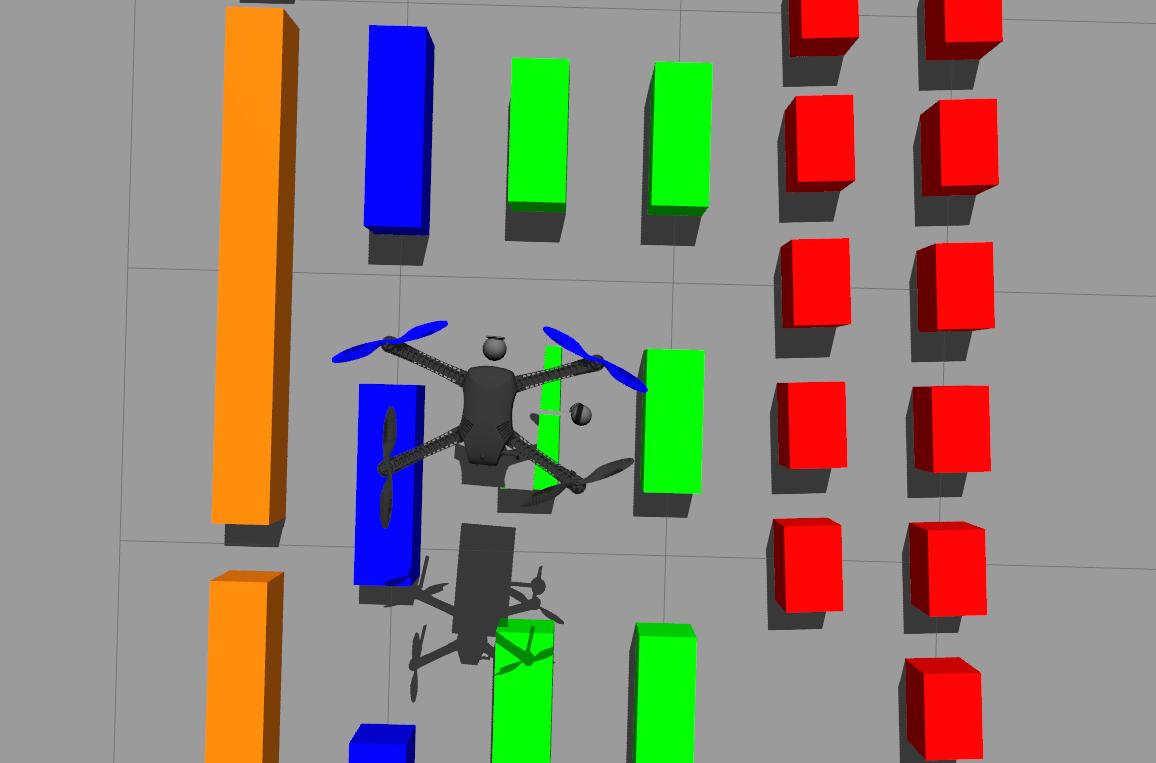} 
\caption{}
\end{subfigure}
\begin{subfigure}{0.33\textwidth}
\includegraphics[scale=0.3]{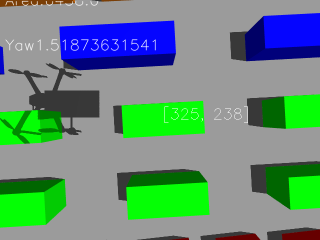} 
\caption{}
\end{subfigure}
\caption{(a) UAV in pick up mode (b) Downward camera frame}
\label{f6}
\end{figure}
\subsubsection{Placing}
Placing involves the following sub-tasks:
\begin{itemize}
\item Channel detection: Near to construction site, the downward facing camera shifted sideways is engaged. The feed is then processed to generate the pose and orientation of the the channel, in case the placement happens directly over the channel or the nearest edge of a brick if placing over a layer. 
\item Drop Pose Update: Based on the pose of the last edge detected, pose and orientation of the brick to be laid is updated.
\item Final Placement: To align the brick with the last laid brick, camera feed from the front cam is processed in a closed loop minimizing the error using a PD controller. The output from the PD controller is then used to update the pose and orientation of manipulator/agent in real time based on the feed. Once the manipulator reaches close to the placing point, the manipulator is disengaged to release the brick.
\begin{figure}
\centering
\includegraphics[scale=0.115]{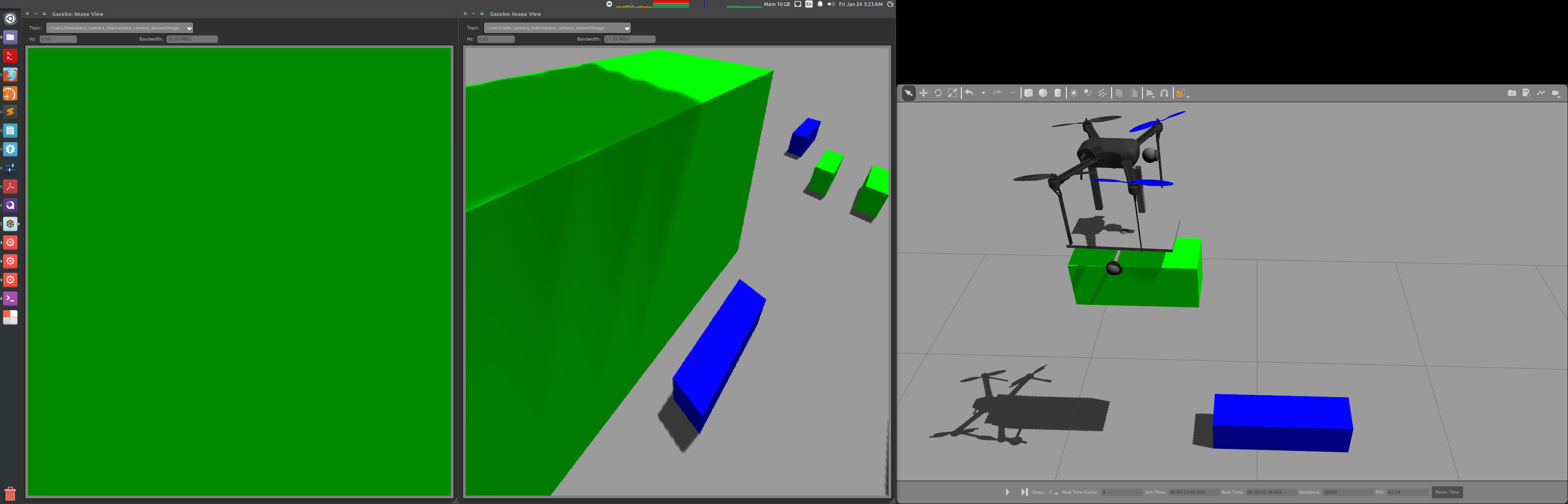}
\caption{Placement operation and camera views in process}
\label{f7}
\end{figure}
\end{itemize}
\section{Construction using UGV}\label{S5}
Similar sub-tasks are involved in pick and place using the ground vehicle. The major areas of difference with respect to the UAV are in the following aspects.
\begin{itemize}
\item Travel (To-Pick): Front camera feed is utilized to detect the center and area of the brick to be picked. Center information of a detected brick is used to orient the UGV and the area information is used to estimate how far the brick is. This information is then fed to a PD control to generate forward velocity and yaw rate for the UGV.
    \[V = kp_v*(d_{area} - b_{area})\]
    \[e_{Cx} = (b_{Cx} - w_{img}/2)\]
    \[\omega_z = kp_z*e_{Cx} + kd_z*(\Delta e_{Cx})\]
where, $V$ is computed velocity of the UGV, $kp_z$ and $kd_z$ are respective proportional and derivative gains, $d_{area}$ is desired area and $b_{area}$ is the captured brick area.
\item Pick Task: The pose and orientation of manipulator arm with respect to robot chassis is used to compute joint angles based on Inverse Kinematics equations.
\begin{itemize}
\item The UGV is guided so as to attain desired $[x, y, z]$ for the manipulator
\item Orientation: For picking operation, the gripper mostly faces vertically down. MOVE-IT package is used for this purpose.
\end{itemize}
\item Travel (to place): The drop location depends upon the picked brick and the closest available drop location. For the UGV to place, the drop position should be in the configurational space of the manipulator.
\item Placing: Upon reaching the placing location, following sub-tasks are initiated.
\begin{itemize}
\item Image Processing: When near to the center but far enough to see the last edge of nearby placed brick, location of the edge is computed in local frame.Based on the image feed of front camera the orientation of the last laid brick with respect to camera is computed.
\item  Drop Pose Update: Based on the pose and orientation of the detected edge, pose and orientation for the brick to be placed is computed.
\item Manipulator Arm Engagement: The manipulator picks the brick back from the carrier, then reaches the position and the orientation computed using inverse kinematics. 
\item Final Placement: A similar  placement procedure is followed as that of UAV.
\end{itemize}
\end{itemize}
\begin{figure}
\centering
\includegraphics[scale=0.5]{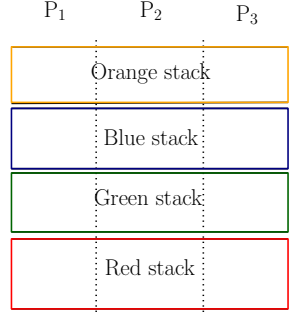}
\caption{The UAV pile division to aid multi-UAV pick without collision. Three partitions are considered here.}
\label{f8}
\end{figure}
\section{Mission Execution}\label{S6}
The sub tasks mentioned in previous section needs to be invoked based on the real time events. A single command evokes the mission planner, which will be running at Ground Control Station (GCS). The external planner needs to have sufficient features not only to take care of the major operations involved but also to handle external events like fail safe, collision avoidance, trial reset, unexpected system reboot, communication loss etc. The features that are available in the developed mission planner are detailed below.
\begin{enumerate}
\item \textbf{Dashboard}: This module is responsible for keeping and updating the complete database of challenge variables. The major data handled are:
\begin{itemize}
\item Pickup Spots: Information of pickup spots including their pose, status, type of brick in it and the points that can be gained from it. This will help to determine free/unhindered spots in the pickup zone.
\item Placement Spots: Information of placements spots including pose, status, layer information (bricks required). This will help determine where to place the picked-up brick and to determine the next required brick to be picked.
\end{itemize}
\item \textbf{Decision Making}: Based on certain cost functions, the agents are chosen which are best suitable for each immediate operation.
\item \textbf{Task Allocation}: After the decision is made, proper task allocation and regular monitoring on the task  is necessary and necessary actions are to be taken if any agent is fails to deliver the task. This is taken care by the task allocation module.
\item \textbf{Fault Handling}: In case of any failure the master needs to be aware of what kind of failure happened and how to ensure further continuity. Some of the failure scenario that can be encountered as:
\begin{itemize}
\item Agent Connectivity Loss
\item Brick Pick Fail
\item Brick Place Fail
\item Collision between the agents
\item Reset/Pause Scenario
\end{itemize}
\item \textbf{Layer Handling}: Given the layer pattern the task is to extract the pose and orientation of each prick in the channel. This information is crucial is decision making process to select from the nearest drop position available for the picked brick.
\item \textbf{Multi-Agent Co-ordination}: This include the following sub parts.
\begin{itemize}
\item Inter-Agent Collision Avoidance: Since multiple agents are involved in the task, there is always a chance of collision. This module ensures enough separation between the agents while travelling, pickup and place. To further reduce the risks of collisions certain constraints are imposed in the strategy.
\item Corridor: The agents are constrained to choose a corridor from three specified heights depending upon the brick they are carrying. The three specified heights to choose from are 3m, 5m and 7m from the ground. Heavier the brick, lower the corridor is chosen to ensure lesser path to travel and lesser turbulence to handle. Even if they happen to cross each other while travelling, the agents will still be separated by a vertical distance of at least 2m.
\end{itemize}
\item \textbf{Blocking occupied channel}: Once the channel is selected for an agent to place, it is blocked for other agents not to choose from. This ensure that the placement task is carried out unhindered.
\item Pickup Channel Scoring: As the pile location is divided into 3 separate arrays with 4 types of brick in each array, there are 12 spots to choose from. This constraint ensures that any pickup spot close to currently picking agent will possess higher cost to select from. Hence this will drive agents to choose most unhindered spot again avoiding collision chances.
\item \textbf{Task Handler}: This module is responsible for allocating tasks along with keeping track of agents and the tasks allocated. Following information is passed into this module.
\begin{itemize}
\item Task Query: For passing the determined tasks to free agent in proper format
\item Task Status: For passing status of the agent and the task allocated
\end{itemize}
Using above two information the handler will be able to allocate task as well as keep track of each of the tasks allocated as whether engaged/completed/failed. The tasks will be allocated according to priority (based on gain/cost) with collision avoidance running in background always to ensure safe operation.
\item \textbf{Task Scheduler}:  This is the core of decision-making system which encodes the strategy chosen to drive the agents to complete the challenge as soon as possible with as much points. When queried by any free agent, this decides the task to be allocated to the agent by determining cost and the gain involved in the overall operation. This is done numerically by defining certain functions and matrices.Cost and gain functions include: 
\begin{itemize}
\item Pickup Cost: This is the cost included in choosing the colour and the spot to pick the chosen colour brick. This helps to determine least hindered place to pick from. Numerically a score matrix of size 4x3 in handling the hindrance involved for UAVs is initialised as
\begin{center}
Score=
$
\begin{bmatrix}
\text 1 & 1 & 1 \\
\text 1 & 1 & 1 \\
\text 1 & 1 & 1 \\
\text 1 & 1 & 1 \\
\end{bmatrix}
$
\end{center}
Upon selection of a particular row and column, say element (2,2), the cost increments happens as:
\begin{center}
Kernel to increase cost:
$
\begin{bmatrix}
1 & 3 & 1 \\
3 & 5 & 3 \\
1 & 3 & 1 \\
1 & 3 & 1
\end{bmatrix}{}
$
Kernel to reset cost: 
$
\begin{bmatrix}
1 & 1/3 & 1 \\
1/3 & 1/5 & 1/3 \\
1 & 1/3 & 1 \\
1 & 1/3 & 1
\end{bmatrix}{}
$
\end{center}

\item Travel Cost: Includes the time taken to travel from pickup spot to building site or vice versa and the height corridor chosen to transport the brick. This will drive the agents to choose the shortest path and the lowest corridor for the heaviest brick. This cost is
    \[Cst_{tr} = k_{tr}*|loc_{targ} - loc_{current}|\]
    where, $Cst$ is the cost incurred, $k_{tr}$ is the gain.
\item Placement Cost: This will help to determine least hindered placement spot and will ensure no other agent enters the channel while one is already placing.
\item Points: This determines the gain achieved in completing a task based on the colour of the brick involved. The gain matrix is
\begin{center}
Points=
$
\begin{bmatrix}
10 & 10 & 10 \\
6 & 6 & 6 \\
4 & 4 & 4 \\
3 & 3 & 3 \\
\end{bmatrix}
$
\end{center}

As shown in the sequence loop, there are two states at which we must decide what to choose.

(a) When Picked: This decision loop will include the placement and the travel cost to determine the best spot to drop the picked brick using the following equation, 
    \[Cst_{drop} = Cst_{placement} + Cst_{tr}\]
(b)) When Dropped/Free: This decision loop will include the travel and pickup cost along with the points gain to determine the colour and spot to choose from using the following equation,
    \[Cst_{pick} = Cst_{spot} + Cst_{tr}\]
\end{itemize}
\end{enumerate}
\section{Simulation Results and Discussion}\label{S7}
The simulating platform is Intel i7 octa-core with 16 GB RAM. This works on Ubuntu 16.04 and ROS kinetic. 
\subsection{UAV operations}
\begin{figure}
    \centering
    \includegraphics[scale=0.15]{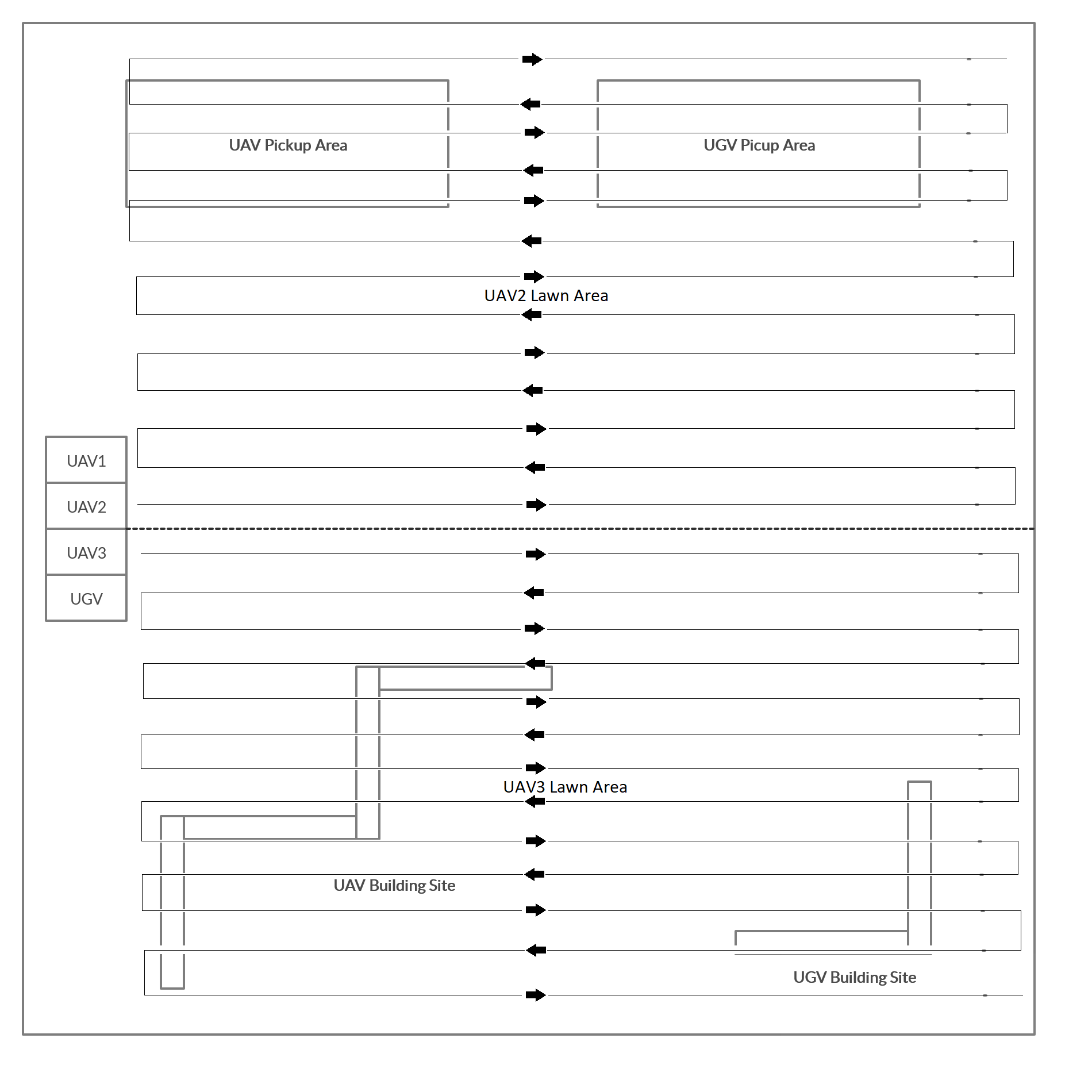}
    \caption{Lawn-mover pattern employed by the UAV to explore the arena}
    \label{f9}
\end{figure}
\begin{figure}
    \centering
\begin{subfigure}{0.45\textwidth}
\includegraphics[scale=0.4]{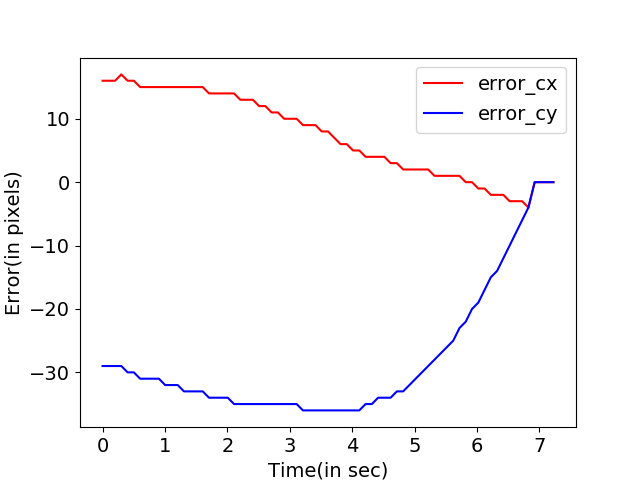} 
\caption{}
\end{subfigure}
\begin{subfigure}{0.45\textwidth}
\includegraphics[scale=0.45]{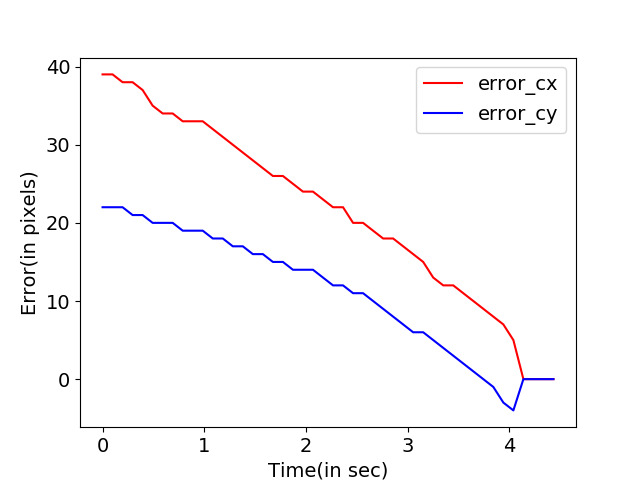} 
\caption{}
\end{subfigure}
    \caption{UAV error profiles in pick and place operations}
    \label{f10}
\end{figure}

Fig. \ref{f9} shows the search pattern followed by the UAVs to detect the brick piles and construction sites for both the UAVs and UGV. The pattern makes sure that the entire arena is explored. The width of this pattern in implementation is decided by the camera FoV. Wider FoV cameras could effectively reduce the number of iterations that the UAV takes to cover the arena. Fig. \ref{f10} gives the error profiles in pixels for picking and placing operations of the UAVs.
As can be seen in Fig. \ref{f10}(a), the PD controller aligns the vehicle and eventually pushes the error to zero for precise pick up of the brick. Fig. \ref{f10}(b) gives a similar profile for placement. 
\begin{figure}
    \centering
    \includegraphics[scale=0.7]{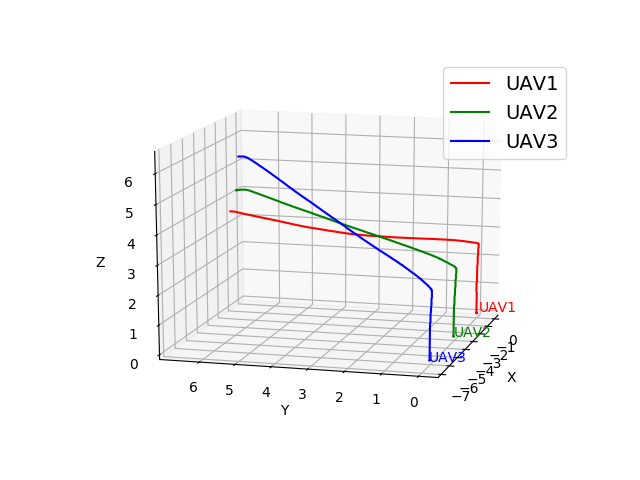}
    \caption{UAV corridors to avoid collision: trajectories}
    \label{f11}
\end{figure}
Fig. \ref{f11} shows the trajectories of the UAVs from take-off until placement of one layer. The drone corridors associated with the architecture makes sure that the vehicles are placed at sufficient separation from each other.

The entire operation process of the UAVs can be found in \footnote{ \href{https://indianinstituteofscience-my.sharepoint.com/:f:/g/personal/kumarankit_iisc_ac_in/EseZ5T-gvDdJqe3Pb5sCyHsB7yxqYbo7VF88ZTQkYUEIIw?e=70KDvi}{UAV operations}}. 
\subsection{UGV operations}
The UGV operations start once the UAV finishes its exploration. The UGV receives the GPS coordinates of its brick stack and starts moving towards the same. Upon detecting different colors, fine corrections are made to direct the vehicle to the specific brick which is decided based on the given sequence. 
\begin{figure}
    \centering
\begin{subfigure}{0.3\textwidth}
\includegraphics[scale=0.265]{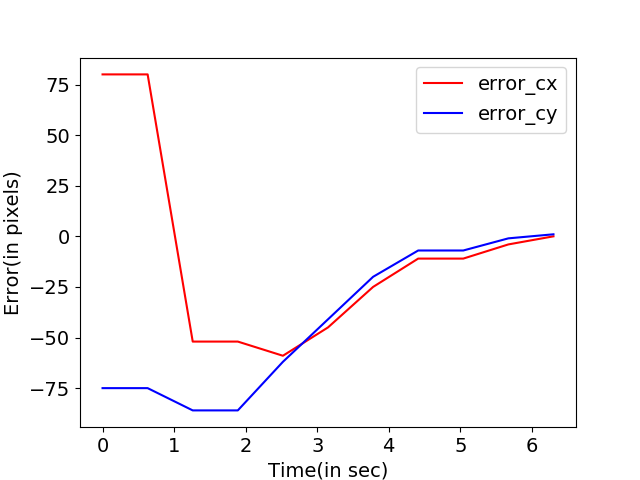} 
\caption{}
\end{subfigure}
\begin{subfigure}{0.33\textwidth}
\includegraphics[scale=0.25]{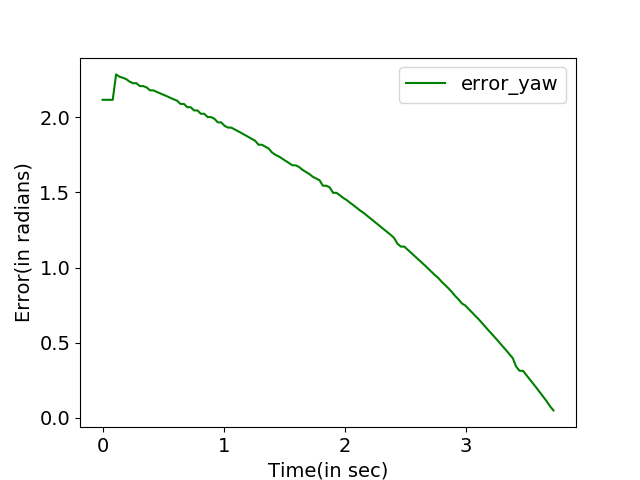} 
\caption{}
\end{subfigure}
\begin{subfigure}{0.3\textwidth}
\includegraphics[scale=0.265]{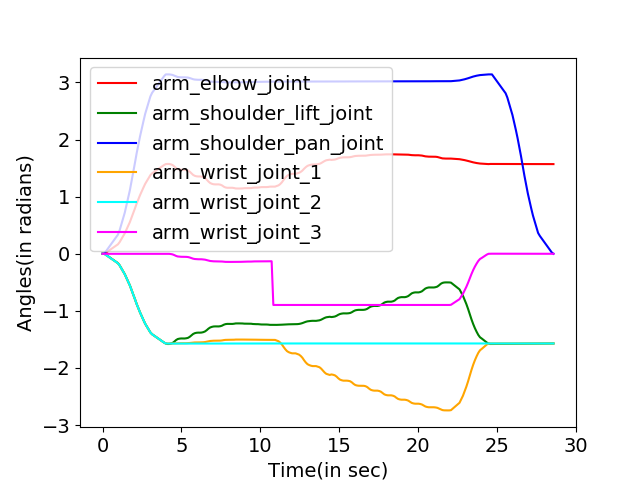} 
\caption{}
\end{subfigure}
    \caption{UGV error profiles in trajectory and heading and joint angle profiles}
    \label{f13}
\end{figure}
Fig. \ref{f13} gives the error plots of UGV trajectory and heading while correcting itself to the brick to be picked up and joint angle profiles in the process of picking up the brick. 

\begin{figure}
    \centering
\begin{subfigure}{0.475\textwidth}
\includegraphics[scale=0.15]{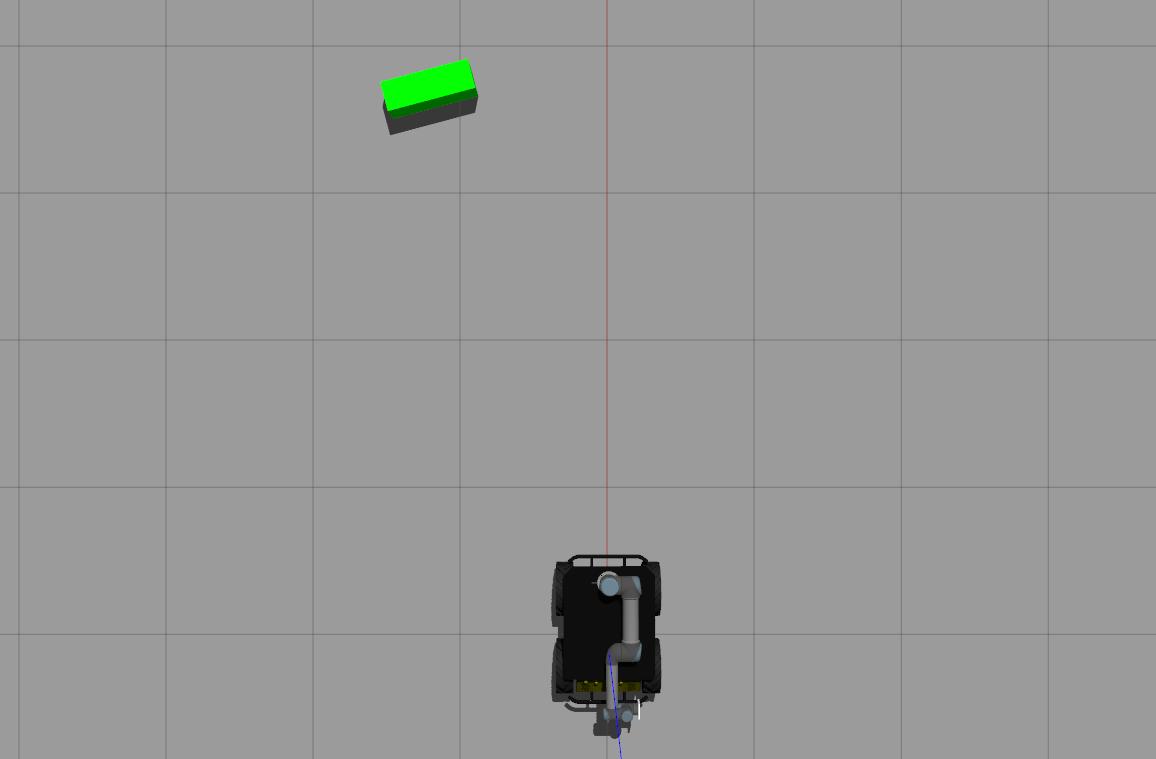} 
\caption{}
\end{subfigure}
\begin{subfigure}{0.5\textwidth}
\includegraphics[scale=0.5]{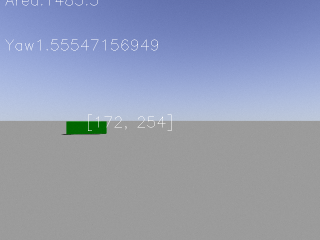} 
\caption{}
\end{subfigure}
\begin{subfigure}{0.475\textwidth}
\includegraphics[scale=0.15]{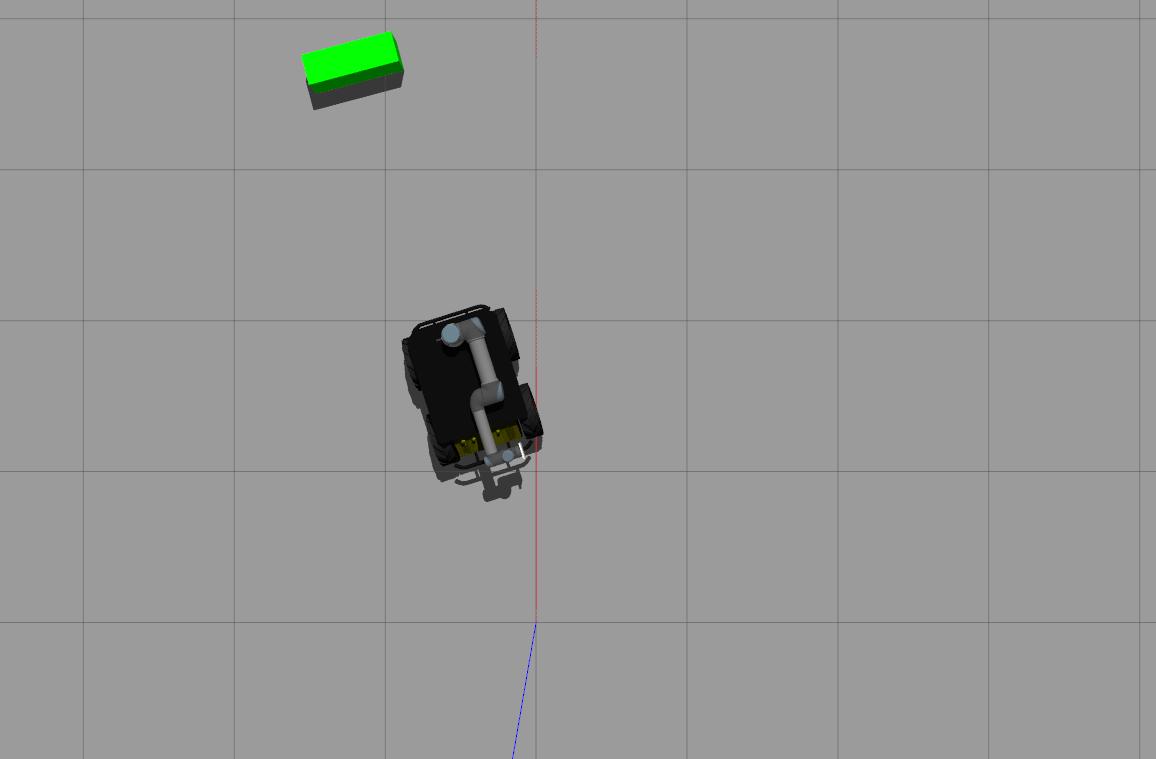} 
\caption{}
\end{subfigure}
\begin{subfigure}{0.5\textwidth}
\includegraphics[scale=0.5]{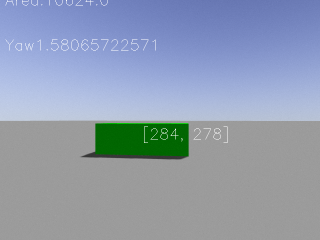} 
\caption{}
\end{subfigure}
    \caption{Snap shots of UGV approaching green brick and respective camera frames denoting the center of detected bricks}
    \label{f15}
\end{figure}
Fig. \ref{f15} shows the snap shots of UGV proceeding to pick a brick. The entire operation process of the UAVs can be found in \footnote{ \href{https://indianinstituteofscience-my.sharepoint.com/:f:/g/personal/kumarankit_iisc_ac_in/Eutq0bj149pMoc_ob511O8MBruChtM5yZTjkWB3NVuJAig?e=v4bUgS}{UGV operations}}.

\section{Discussion}
This work is intended to be a reference material for an end-to-end development of software flight stack and peripheral task planner for building a wall using multiple UAVs and UGV. This includes control, guidance, vision and manipulation aspects of the entire system. Nevertheless, there are certain aspects of improvement which could be appended to this work to make it more innovative. One among this is UAV-UGV Collaboration. In the current work, UAVs collaborate among themselves to build the structure while UGV build its structure on its own. But, as UAVs could finish the work faster compared to the UGV, the UAVs could assist in speeding up the UGV wall building. Certain aspects which would be appended in the future extensions of this work would be
\begin{itemize}
\item Intermediate Drop-Pickup Zone: This type of collaboration is helpful in UGV construction site. Since the agility of UGV is very less compared to the UAVs, UAVs could be deployed to carry the bricks from the UGV pile location to an intermediate drop location near to UGV construction site. The bricks carried by the UAVs will be placed in proper pose and orientation nearby so as the UGV need not traverse much in collecting bricks for construction. This will overall improve the speed of layer building for UGV.
\item Heavy/In Bulk Brick Carriage: This strategy involves UGV to carry brick in bulk amount (at least 3 or depending upon the agent requirements) from the UAV pile location and drop near to UAV construction site. This will greatly reduce the time UAVs spent carrying the bricks in the turbulent environment. This is especially applicable for orange brick which is the longest one and which causes issues for the UAVs to carry on its own.
\end{itemize}
Multi-UAV collaboration to pick longest brick is another aspect which is to be addressed. This capability would accelerate the UAV wall building where the orange bricks could be placed with better precision and ease.
\section{Conclusions}\label{S8}
This paper is motivated by MBZIRC 2020 challenge 2 which involves picking and placing of objects of different sizes and weights so as to build a structure. The complete software framework involved is described in this which details each and every aspect required for the successful execution of the challenge. This includes the operations of both UAV as well as UGV. The complete pipeline required for the challenge is developed which uses classical methods to carry out the challenge in ROS-Gazebo framework. Simulation results verify this ready to implement framework. Implementation on real hardware is under progress and would be reported in near future publications.
\begin{acknowledgements}
We would like to acknowledge the Robert Bosch Center for Cyber Physical Systems, Indian Institute of Science, Bangalore, and Khalifa University, Abu Dhabi, for partial financial support. We would also like to thank fellow team members from IISc for their invaluable contributions towards this competition.
\end{acknowledgements}

\end{document}